\documentclass[letterpaper, 10 pt, conference]{ieeeconf} 
\IEEEoverridecommandlockouts
\usepackage[utf8]{inputenc}
\usepackage[T1]{fontenc}
\usepackage{comment}
\usepackage{graphicx} 
\usepackage{dblfloatfix} 
\graphicspath{ {Figures/} }
\usepackage{amsmath, amsfonts, amssymb, bm}
\usepackage[usenames,dvipsnames]{xcolor}
\usepackage{array}
\usepackage{setspace}
\newcolumntype{L}{>{\centering\arraybackslash}m{0.45in}}
\newcolumntype{W}{>{\centering\arraybackslash}m{0.8in}}
\usepackage{threeparttable}
\usepackage{bm}

\usepackage{float}
\usepackage{caption}
\usepackage{subcaption}
\usepackage{accents}
\usepackage{hyperref}
\usepackage{cite}

\usepackage[labelfont=bf, font=small]{caption}

\let\labelindent\relax
\usepackage{enumitem}

\DeclareSymbolFont{symbolsC}{U}{txsyc}{m}{n}
\DeclareMathSymbol{\circleleft}{\mathrel}{symbolsC}{146}
\DeclareMathSymbol{\circleright}{\mathrel}{symbolsC}{145}

\DeclareRobustCommand{\rightgroupderiv}[1]{\accentset{\scriptscriptstyle{\circleright}}{#1}}

\newcommand{\Adjointinv}[1]{\textrm{Ad}_{#1}^{\textrm{-}1}}

\title{\LARGE \bf A Fast and Model Based Approach for Evaluating Task-Competence

of Antagonistic Continuum Arms}

\author{Bill Fan$^{1}$, Jacob Roulier$^{2}$, and Gina Olson$^{2}$%
\thanks{$^{1}$ Olin College of Engineering, Needham, MA, USA.}
\thanks{$^{2}$Department of Mechanical and Industrial Engineering, University of Massachusetts, Amherst, MA, USA.}%
}

\begin{document}

\maketitle
\thispagestyle{empty}
\pagestyle{empty}

\begin{abstract}
Soft robot arms have made significant progress towards completing human-scale tasks, but designing arms for tasks with specific load and workspace requirements remains difficult. 
A key challenge is the lack of model-based design tools, forcing advancement to occur through empirical iteration and observation. 
Existing models are focused on control and rely on parameter fits, which means they cannot provide general conclusions about the mapping between design and performance or the influence of factors outside the fitting data.
As a first step toward model-based design tools, we introduce a novel method of analyzing whether a proposed arm design can complete desired tasks. Our method is informative, interpretable, and fast; it provides novel metrics for quantifying a proposed arm design's ability to perform a task, it yields a graphical interpretation of performance through segment forces, and computing it is over 80x faster than optimization based methods.
Our formulation focuses on antagonistic, pneumatically-driven soft arms.
We demonstrate our approach through example analysis, and also through consideration of antagonistic vs non-antagonistic designs. 
Our method enables fast, direct and task-specific comparison of these two architectures, and provides a new visualization of the comparative mechanics. 
While only a first step, the proposed approach will support advancement of model-based design tools, leading to highly capable soft arms.

\end{abstract}

\section{INTRODUCTION}

\begin{figure}
    \centering
    \includegraphics[width=\linewidth]{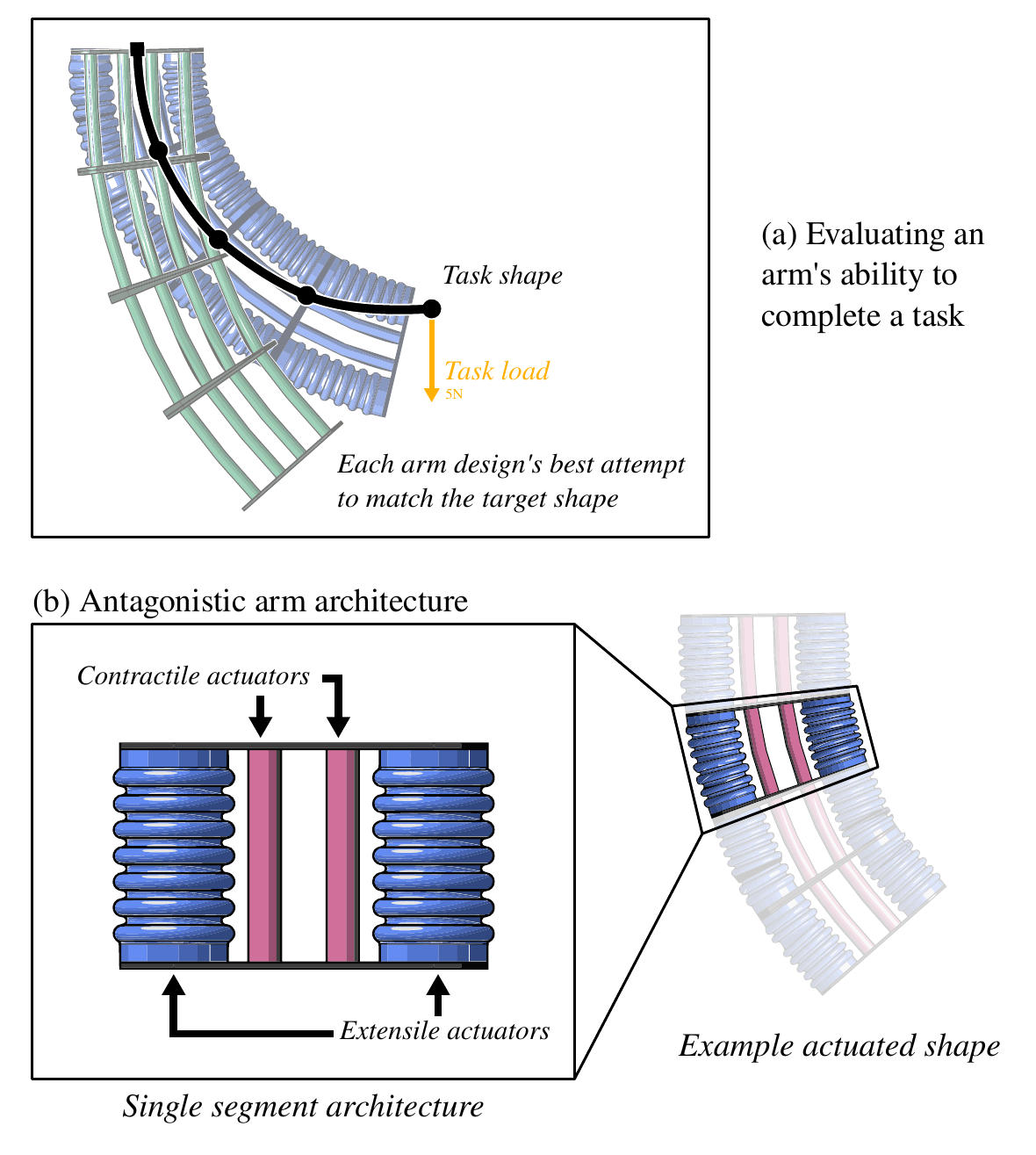}
    \caption{\textit{(A):} We propose a method to evaluate fluid-driven soft arm designs on their ability to sustain a task shape at the task load. \textit{(B):} We mainly demonstrate our method with planar antagonistic arm designs. Each arm segment has two extending and two contracting actuators, and arms move via selective actuator pressurization.}
    \label{fig:intro_fig}
    \vspace{-1.5em}
\end{figure}

Fluid-driven soft robot arms seek to capture the physical intelligence of muscular hydrostats, such as elephant trunks and octopus arms \cite{kier_1985}, in order to improve robot robustness and safety around humans \cite{chen_review_2022}. 
These arms combine soft pneumatic actuators in parallel and in series to produce robotic arms that can bend in any direction at multiple points (see example in Fig. \ref{fig:intro_fig}) \cite{octarm,davis_bidirectional_varstiff,stiff_flop_jamming,jiang_hierarchical_2021}.
Recent works have taken major steps toward soft robotic arms that can complete human-scale tasks, exploring backbone-free antagonistic designs for stiffness control \cite{var_grip_stiffness_independence,davis_varstiff_advancedrobotics,OlsonAsselmeier_RoboSoft} and demonstrating contact rich tasks such as opening drawers \cite{jiang_hierarchical_2021}, washing human subjects \cite{zlatintsi_i-support_2020} and assisted eating \cite{guan_trimmed_2023}. However, arm demonstrations are consistently limited to small external loads, even with high actuator pressures \cite{octarm,davis_varstiff_advancedrobotics,jiang_design_2016,jiang_hierarchical_2021,guan_trimmed_2023}.

%
%

The mechanical reason for limited force is not obvious, but it is evident even in recent, novel demonstrations: in Jiang et al., no tasks are demonstrated with additional weights \cite{jiang_hierarchical_2021}, and in related work the arm's maximum characterized load is 2.8N \cite{jiang_hierarchical_2021}\cite{jiang_design_2016}. The arm in Guan et al. is shown, in a separate work, to deform its entire body length under 4N of tip loading \cite{guan_trimmed_2023, stella_piecewise_2023_s}. Investigations of the mapping between design and performance, e.g., load capacity across the workspace, are hampered by the lack of model-based design tools for soft robotic arm. Soft arm modeling efforts have focused on control, and use experimental, homogenized stiffness parameters that are specific to the design being tested \cite{renda_cosserat_tro,trivedi_model_tro,rus_ijrr_control}. Practically, an arm's ability to complete a desired task has been determined by building and testing it. 

%
%

%
%
%

\begin{figure*}[!ht]
    \centering  
    \includegraphics[width=\textwidth]{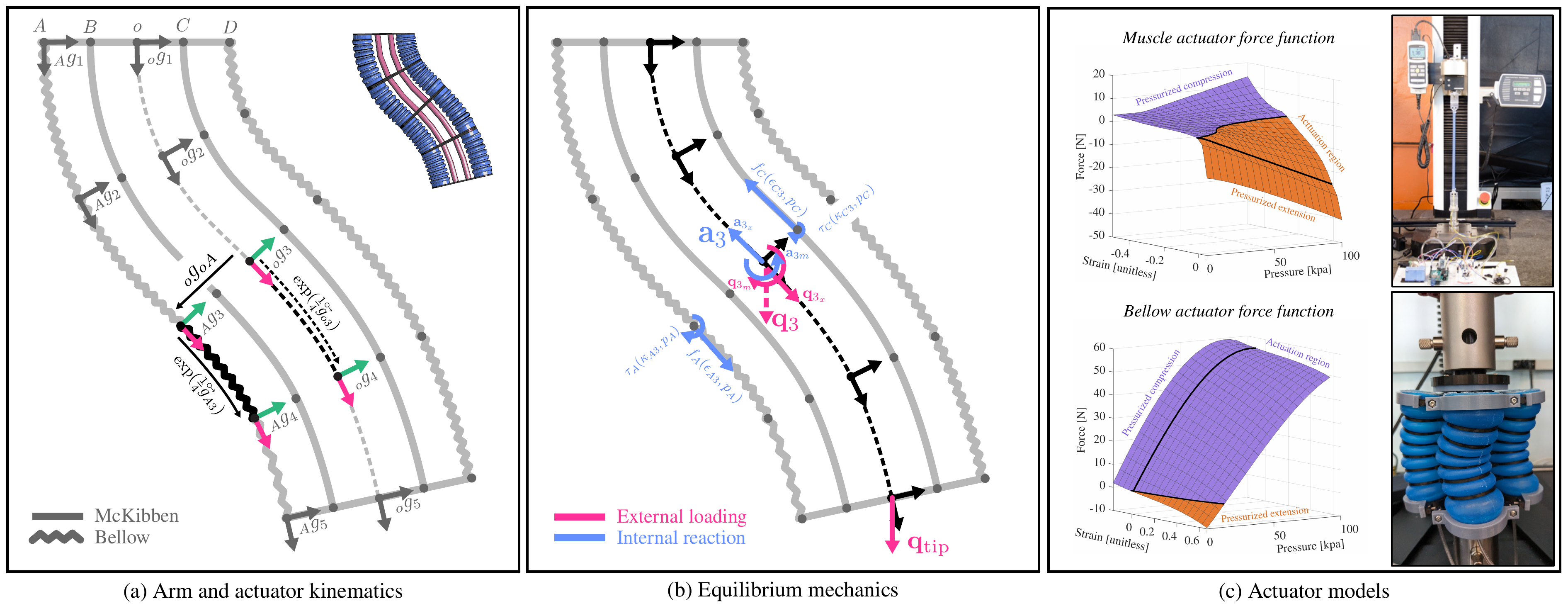}
    \caption{\textbf{Overview of our mechanics model}. \textit{(A):} A soft robot arm composed of two bellows and two muscles can be fully parameterized by its centerline twists $\rightgroupderiv{g}_{oi}$ and transformations between actuators ${}_og_{o\alpha}$. \textit{(B):} When a tip load $\mathbf{q}_{\textrm{tip}}$ is applied, wrenches $\mathbf{q}_i$ are induced along the arm's backbone. To achieve static equilibrium, each actuator contributes a reaction force and moment to balance the load. \textit{(C):}   We consider two types of actuators: contracting McKibben artificial muscles and extensile bellows actuators. The characterized force functions $f(\epsilon, p)$ of each actuator are also shown, with actuation regimes labeled according to \cite{olson_eulerbernoulli_2020}.}
    \label{fig:model_explanation}
    \vspace{-1em}
\end{figure*}

Model-based design tools require model formulations that generalize across designs and methods of using those models to produce informative, interpretable results. Prior work \cite{olson_eulerbernoulli_2020} has developed generalizable models of soft arms, but significant gaps remain in developing approaches that utilize these models to provide insights about the mapping between arm design and task-specific performance. 

In this work, we develop a novel, model-based method for evaluating a proposed design's ability to complete specific tasks. 
Our method is informative, interpretable, fast and provides a visualization of segment capabilities. We use it to concretely establish that antagonistic arms can complete a wider range of tasks than non-antagonistic arms, we provide novel insights for why, and computing it is 80x faster than existing methods. 
We first establish our underlying mechanics model in Section \ref{sec:model}, then introduce our method for analyzing task attainability in Section \ref{sec:attainability}, and finally apply our method to the comparison of arm designs in Section \ref{sec:results}. 
\section{CONTINUUM MODEL} 
\label{sec:model}

This section briefly reviews our underlying assumptions, notation, and mechanics-based continuum model, which is a planar Cosserat-rod formulation of the design-oriented model of \cite{olson_eulerbernoulli_2020}. Our work makes the following assumptions:

\begin{enumerate}[label=\textbf{\arabic*})]
    \item{\textbf{Design:} We assume the arm and actuators act as quasi-static shear-free, torsion-free rods. Prior work has shown that while this may not be valid in minimal cases (three long, thin actuators), it becomes reasonable when stiffer, shorter or more actuators are used \cite{me_mocap}.}
    \item{\textbf{Actuators:} The arm's active elements are fluid-driven soft actuators, with uniaxial force dependent on strain and pressure. We assume for any constant strain the actuator's pressure-force function is continuous and monotonic. The actuators are discretized into multiple segments for numerical solution but share pressures.}
    \item{\textbf{Bending Stiffness:} We assume the actuators have bending stiffness beyond that generated their uniaxial forces, and is dependent on curvature and pressure.}
    \item{\textbf{Generality:} For clarity, we develop the model and consider examples for planar arms with four actuators and end loads, but our methods can be adapted to include more actuators, 3D configurations and distributed loads.}
\end{enumerate}

See our SI for the forms of all matrices used below. \footnote{For our SI, proofs, code, and figures, see \url{https://github.com/wfan19/antagonistic-task-competency}}


\subsection{Cosserat rod kinematics} \label{ssec:cosserat_kinematics}

Consider a robot arm with $M$ actuators $\{A, B, \dots, \mu\}$ discretized by $N+1$ nodes into $N$ constant curvature segments ($\mu$ stands for the $M$-th actuator). Note that discretization is for numerical solution, and does not represent physical segmentation. Let $\alpha$ be any actuator. Each actuator $\alpha$ is mounted at a distance of $r_\alpha$ away from the center. For any actuator $\alpha$, the pose of its $i$-th node can be denoted as ${}_{\alpha}g_{i} = \begin{bmatrix} {}_{\alpha}x_{i} & {}_{\alpha}y_{i} & {}_{\alpha}\theta_{i} \end{bmatrix}^\textrm{T}$, with matrix representation:

\begin{equation}
    \rho({}_{\alpha}g_{i}) = 
    \begin{bmatrix}
        \cos({}_{\alpha} \theta_{i}) & -\sin({}_{\alpha} \theta_{i}) & {}_{\alpha}x_{i} \\
        \sin({}_{\alpha} \theta_{i}) & \cos({}_{\alpha} \theta_{i}) & {}_{\alpha}y_{i} \\
        0 & 0 & 1
    \end{bmatrix}
\end{equation}

We will notate pose composition as ${}_{\alpha}g_2 = {}_{\alpha}g_1 \circ {}_{\alpha}g_{12}$, which is implemented as the product of the matrix forms $\rho({}_{\alpha}g_2) = \rho({}_{\alpha}g_1) \rho({}_{\alpha}g_{12})$. At each node $i$ of actuator $\alpha$, the twist vector ${}_{\alpha}\rightgroupderiv{g}_{i} = \begin{bmatrix} {}_{\alpha}l_{i} & {}_{\alpha}\gamma_{i} & {}_{\alpha}\kappa_{i}\end{bmatrix}^\textrm{T}$ describes the body-frame rate of change of $g_{\alpha i}$ with $l$ as the instantaneous length of the segment, $\gamma$ the shear, and $\kappa$ the curvature. The matrix representation of the twist vector is:
\begin{equation}
    \rho({}_{\alpha}\rightgroupderiv{g}_{i}) =
    \begin{bmatrix}
        0 & {}_{\alpha}\kappa_{i} & {}_{\alpha}l_{i} \\
        {}_{\alpha}\kappa_{i} & 0 & {}_{\alpha}\gamma_{i} \\
        0 & 0 & 0
    \end{bmatrix}
\end{equation}

These twists can be integrated using the exponential map $\exp(\rightgroupderiv{g}) = \exp_M(\rho(\rightgroupderiv{g}))$ to recover poses along each actuator, where $\exp_M$ is the matrix exponential:

\begin{equation} \label{eqn:product_of_exponentials}
    {}_{\alpha}g_{i} = {}_{\alpha}g_{1} \circ \prod_{k=1}^{i-1}\exp(\frac{1}{n} \rightgroupderiv{{}_{\alpha}g_k})
\end{equation}

Using eqn. \ref{eqn:product_of_exponentials}, we can describe the poses along each actuator ${}_{\alpha}g = \{{}_{\alpha}g_{1}, {}_{\alpha}g_{2}, \cdots {}_{\alpha}g_{N+1}\}$ using the base pose ${}_{\alpha}g_{1}$ and the twists of each segment ${}_{\alpha}\rightgroupderiv{g} = \{{}_{\alpha}\rightgroupderiv{g}_{1}, {}_{\alpha}\rightgroupderiv{g}_{2}, \cdots, {}_{\alpha}\rightgroupderiv{g}_{N}\}$, which are more closely related to the mechanics (Fig. \ref{fig:model_explanation}A). Furthermore, our assumption of a rigid and constant cross-section enables us relate the twists of each actuator to the twists of the manipulator centerline. Let the pose and twists of the manipulator centerline at node $i$ be ${}_{o}g_i$ and ${}_{o}\rightgroupderiv{g}_i$ respectively, and let ${}_{o}g_{o\alpha}$ be the transformation from centerline $o$ to actuator $\alpha$ in each cross-section, which is usually $\begin{bmatrix}0 & r_\alpha & 0\end{bmatrix}^\textrm{T}$. As shown in \cite{fan_linear_2023}, the twist of any actuator can be computed from the centerline twist as:

\begin{equation}
    {}_{\alpha}\rightgroupderiv{g}_{i} = \Adjointinv{o\alpha} {}_{o}\rightgroupderiv{g}_{i}
\end{equation}

\begin{figure*}[!ht]
    \centering
    \includegraphics[width=\textwidth]{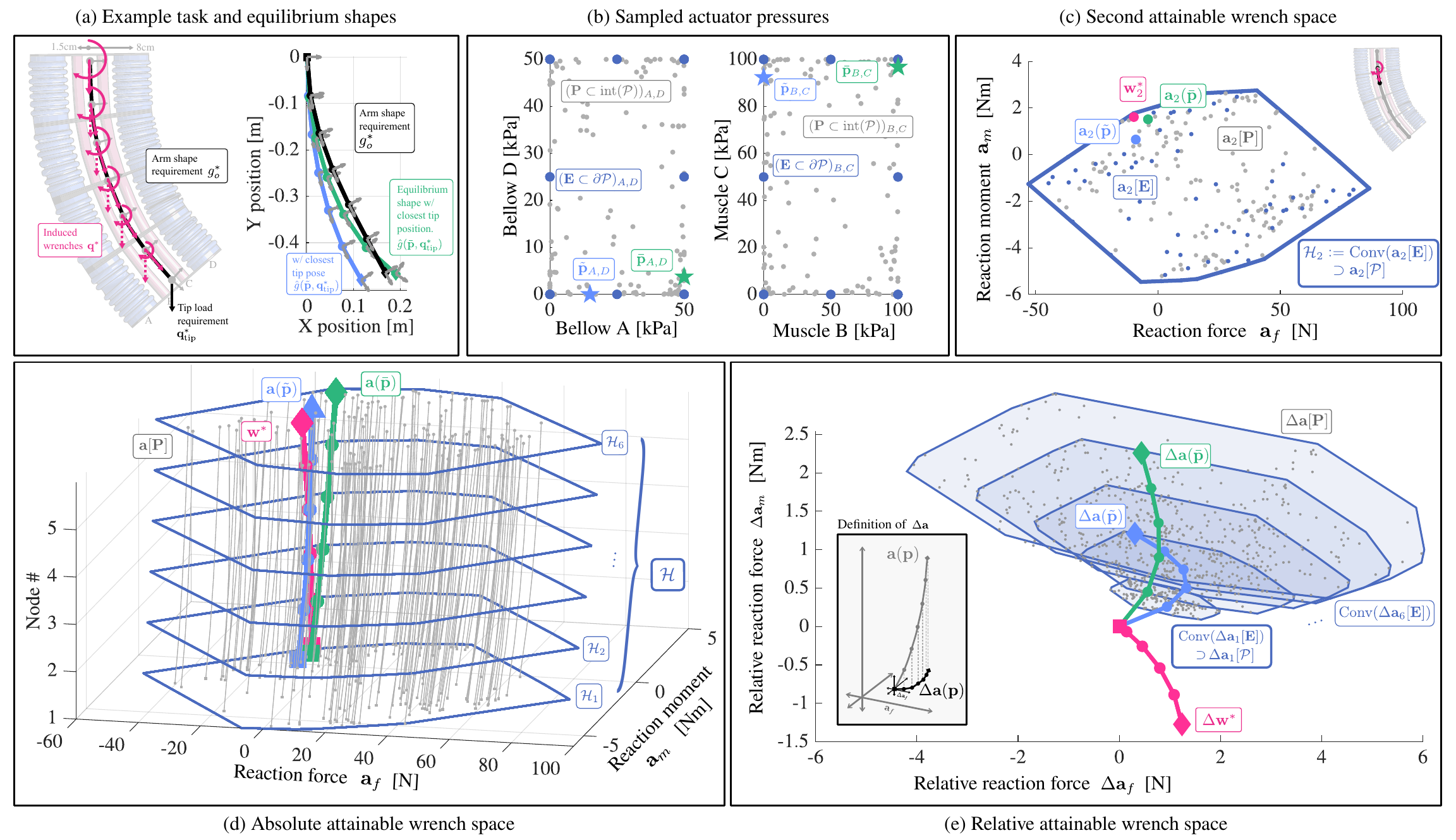}
        \caption{\textbf{Visual explanation of attainable wrench spaces}. \textit{(A-left):} an example task shape and load, and consequent requirement wrenches. \textit{(A-right):} equilibrium shapes that minimize tip position error or tip pose error compared to the specified task shape. \textit{(B):} Points are uniformly sampled from the pressure space's edges uniformly, and from the interior using a beta distribution with $\alpha = \beta = 0.3$. Stars are the pressures $\tilde{\mathbf{p}}$ and $\bar{\mathbf{p}}$ which yield equilibrium shapes in \textit{(A-right)}. \textit{(C):} The 2nd attainable wrench hull $\mathcal{H}_2$, with reaction and requirement wrenches superimposed. \textit{(D):} Attainable wrench sequences corresponding to sampled pressures in \textit{(B)} and solution pressures from \textit{(A)}. \textit{(E):} Relative wrench sequences - the lines of the relative attainable wrench sequences are omitted for clarity.}
    \centering
    \label{fig:wrench_hull_explanation}
    \vspace{-1.5em}
\end{figure*}

Thus, an arm's configuration can be fully parameterized by its centerline base-pose ${}_{o}g_1$, centerline twists ${}_{o}\rightgroupderiv{g} = \{ {}_{o}\rightgroupderiv{g}_1, {}_{o}\rightgroupderiv{g}_2, \dots, {}_{o}\rightgroupderiv{g}_N \}$, and the position of each actuator $r_\alpha$.

\subsection{External Loads}
Static equilibrium is achieved by balancing internal reactions to external loading. We will refer to the combination of a force and moment at a point as a wrench, and consider external wrenches applied to the tip of a robot arm $\mathbf{q}_{\textrm{tip}} = \begin{bmatrix}f_x & f_y & m\end{bmatrix}^\textrm{T}$. This tip wrench will induce a wrench at every node along the arm's centerline with equal force but additional moment due to the moment arm (see Fig. \ref{fig:wrench_hull_explanation}A-left). The wrench at each node $i$ along the arm's centerline is:

\begin{equation}
    \mathbf{q}_i = {}_{i}\textrm{Ad}^{-\textrm{T}}_{in} T_eL_{g_n}^\textrm{T} \mathbf{q}_{\textrm{tip}}
\end{equation}

Note $\mathbf{q}_i$ is driven by centerline shape and external loads, while independent of actuation, arm design, or materials.

\subsection{Internal Reaction}

Given the resting length of an actuator $\alpha$ as ${}_{\alpha}\hat{l}$, the strain ${}_{\alpha}\epsilon_i$ of $\alpha$ at node $i$ can be computed using the length component of the twist-vector ${}_{\alpha}\rightgroupderiv{g}_{i}$:

\begin{equation}
    {}_{\alpha}\epsilon_i = ({}_{\alpha}l_{i} - {}_{\alpha}\hat{l}) / {}_{\alpha}\hat{l}
\end{equation}

Actuator deformation from resting length and any curvature will result in reaction forces and moments. We characterize each actuator's force as a function of strain and pressure $f(\epsilon, p)$, and its bending moment as a function of curvature and pressure $\tau(\kappa, p)$. Each actuator's reaction force and moment contributes to an overall reaction wrench along the arm's centerline, which can be found as follows:


{
\thinmuskip=1mu
\begin{align}
    &\mathbf{a}_i
    = \; \begin{bmatrix}
        1 & 1 & \dots & 1 \\
        0 & 0 & \dots & 0 \\
        r_1 & r_2 & \dots & r_m
    \end{bmatrix}
    \begin{bmatrix}
        f_A({}_{A}\epsilon_i, p_A) \\
        f_B({}_{B}\epsilon_i, p_B) \\
        \vdots \\
        f_\mu({}_{\mu}\epsilon_i, p_\mu)
    \end{bmatrix}
    +
    \\
    &1e3\begin{bmatrix}
        0 & 0 & \dots & 0 \\
        1 & 1 & \dots & 1 \\
        0 & 0 & \dots & 0
    \end{bmatrix}
    \begin{bmatrix}
        {}_{A}\gamma_i \\
        {}_{B}\gamma_i \\
        \vdots \\
        {}_{\mu}\gamma_i
    \end{bmatrix}
    + \begin{bmatrix}
        0 & 0 & \dots & 0 \\
        0 & 0 & \dots & 0 \\
        1 & 1 & \dots & 1
    \end{bmatrix}
    \begin{bmatrix}
        \tau_A({}_{A}\kappa_i, p_A) \\
        \tau_B({}_{B}\kappa_i, p_B) \\
        \vdots \\
        \tau_\mu({}_{\mu}\kappa_i, p_\mu)
    \end{bmatrix}\nonumber
\end{align}
}

where the $\gamma$ term simulates the arm's high shear stiffness.

\subsection{Equilibrium model} \label{ssec:equilibrium_model}

The above equations can be combined into the governing equation of a soft robot arm. Fixing the actuator positions $\mathbf{r} = \{r_A, r_B, \dots, r_\mu \}$, neutral lengths $\mathbf{\hat{l}} = \{\hat{l}_A, \hat{l}_B, \dots, \hat{l}_\mu \}$, and actuator characteristics $\mathcal{C} = \{f_A, \tau_A, f_B, \tau_B, \dots, f_\mu, \tau_\mu \}$ of an arm - which we collectively call its \textit{\textbf{design}} - its equilibrium shape ${}_{o}\hat{\rightgroupderiv{g}}$ when subjected to tip load $\mathbf{q}_{\textrm{tip}}$ and actuated to pressure $\mathbf{p}$ is found by solving the following equation for ${}_{o}\hat{\rightgroupderiv{g}}$ at each $i$:

\begin{equation} \label{eqn:separated_equilibrium_eqn}
    \mathbf{a}_i({}_{o}\hat{\rightgroupderiv{g}}_{i}, \mathbf{p}) + \mathbf{q}_i( {}_{o}\hat{\rightgroupderiv{g}}, \mathbf{q}_{\textrm{tip}}) = \mathbf{0}
\end{equation}

We will refer to the act of solving the above equation for the equilibrium twists ${}_{o}\hat{\rightgroupderiv{g}}$ when given a pressure $\mathbf{p}$ and tip load $\mathbf{q}_{\textrm{tip}}$ as solving the "forward mechanics", and notate it as the mapping ${}_{o}\hat{\rightgroupderiv{g}}(\mathbf{p}, \mathbf{q}_{\textrm{tip}})$, whose poses are ${}_{o}\hat{g}(\mathbf{p}, \mathbf{q}_{\textrm{tip}})$.

\section{TASK ATTAINABILITY ANALYSIS}
\label{sec:attainability}

We will now use our mechanics model to evaluate whether a proposed arm design can accomplish desired tasks. We begin by stating the problem and defining relevant concepts, and then introducing a search-based solution, which is comprehensive but slow. Then, we will introduce our approximation of the problem as a convex quadratic problem, which is not only faster, but also more interpretable.

\subsection{Problem Statement}
Given an arm design as defined above, we are interested in its ability to sustain desired shapes when carrying specific payloads. Specifically, when an arm is subject to a tip load $\mathbf{q}_{\textrm{tip}}^*$, we ask whether it can maintain a desired equilibrium centerline shape ${}_{o}g^* = \{{}_{o}g^*_1, {}_{o}g^*_2, \dots, {}_{o}g^*_{N+1}\}$, or equivalently twists ${}_{o}\rightgroupderiv{g}^* = \{{}_{o}\rightgroupderiv{g}^*_1, {}_{o}\rightgroupderiv{g}^*_2, \dots, {}_{o}\rightgroupderiv{g}^*_N \}$. We define the pair $(\mathbf{q}_{\textrm{tip}}^*, {}_{o}\rightgroupderiv{g}^*)$ as a \textbf{\textit{task}}, and refer to its components respectively as the \textit{\textbf{task load}} and \textit{\textbf{task shape}}.

An arm's control space is defined by the pressure limits of each actuator. If each actuator has max pressure $\bar{p}_A, \bar{p}_B, \dots, \bar{p}_\mu$ and minimum pressure zero, then the input pressure space is $\mathcal{P} = [0, \bar{p}_A] \times [0, \bar{p}_B] \times \dots \times [0, \bar{p}_\mu]$. When a proposed arm design is capable of completing a desired task, this means that there exists a pressure $\mathbf{p}^*$ in $\mathcal{P}$ such that the arm's equilibrium shape when subject to the task load is equal to the specified task shape, i.e. ${}_{o}\hat{g}(\mathbf{p}^*, \mathbf{q}_\textrm{tip}^*) = {}_{o}g^*$. If this is true, we say the task is \textit{\textbf{attainable}}.

\subsection{Solution through search} \label{sec:naive_method}

The most immediate way to determine an arm's ability to attain a task is to search the pressure space $\mathcal{P}$ for a pressure $\mathbf{p}$ whose actuated equilibrium shape ${}_{o}\hat{g}(\mathbf{p}, \mathbf{q}^*_{\textrm{tip}})$ most closely matches the task-shape ${}_{o}g^*$. This can be solved as:


\begin{equation} \label{eqn:naive_attainability}
    s = \min_{\mathbf{p} \in \mathcal{P}} \sum_{i = 1}^N ({}_{o}\hat{g}_i(\mathbf{p})^{-1} \circ {}_{o}g_i^*)^\textrm{T} \mathbf{K}_i ({}_{o}\hat{g}_i(\mathbf{p})^{-1} \circ {}_{o}g_i^*)
\end{equation}

The solution $s$ is the summed difference between the poses of the task-shape and its closest possible equilibrium shape. If $s$ is zero, then there exists $\mathbf{p}$ in $\mathcal{P}$ such that ${}_{o}\hat{g}(\mathbf{p}) = {}_{o}g^*$, and the task is attainable. This method is exhaustive and flexible - each node's position errors can be weighed against angular ones using $\mathbf{K}_i$, or ignored entirely (see Fig \ref{fig:wrench_hull_explanation}A-right). However, this method is  slow, as each step of the optimization requires a solution of the forward mechanics, and the optimization itself is nonlinear and non-convex.

\subsection{Wrench-Hull Analysis}
To simplify our problem, we will consider our requirements in wrench-space rather than Euclidean space, which eliminates the need to solve the forward mechanics and reveals interesting arm properties. First, notice that specifying a task $({}_{o}\hat{\rightgroupderiv{g}}, \mathbf{q}_{\textrm{tip}})$ makes eqn. \ref{eqn:separated_equilibrium_eqn} solely dependent on pressure:

\begin{equation} \label{eqn:rxn_target}
    \mathbf{a}_i(\mathbf{p}) = -\mathbf{q}_i^*
\end{equation}

In this case, specifying a task defines both load and desired arm shape, which is more restrictive than specifying only desired tip position.
For the specified task to be attainable, eqn. \ref{eqn:rxn_target} must be true across all $i$. Let us denote $\mathbf{a}(\mathbf{p}) = (\mathbf{a}_1(\mathbf{p}), \dots, \mathbf{a}_N(\mathbf{p}))$ an \textbf{\textit{attainable wrench sequence}}, and $\mathbf{w}^* = -\mathbf{q}^* = (-\mathbf{q}^*_1, \dots, -\mathbf{q}^*_N)$ a \textbf{\textit{requirement wrench sequence}}. Thus, we want to determine whether there exists a $\mathbf{p}$ in $\mathcal{P}$ such that $\mathbf{a}(\mathbf{p}) = \mathbf{w}^*$. Crucially, this is equivalent to determining whether $\mathbf{w}^*$ lies within $\mathbf{a}[\mathcal{P}] = \{ \mathbf{a}(\mathbf{p}) \mid \mathbf{p} \in \mathcal{P} \}$, which is the space of attainable wrench sequences.

\begin{figure*}[!ht]
    \centering
    \includegraphics[width=\textwidth]{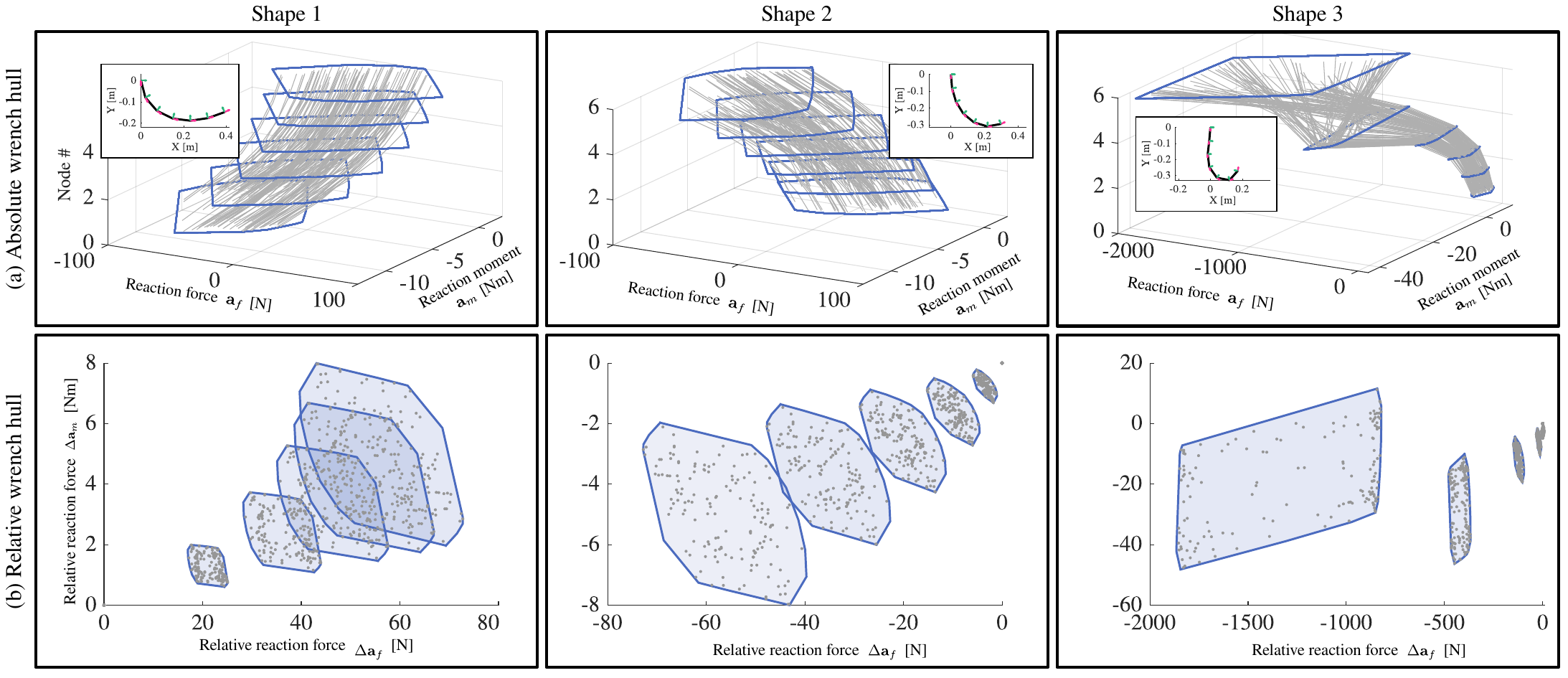}
    \caption{\textbf{Visualization of wrench-spaces for different arm shapes}. Using the arm design from Fig. \ref{fig:wrench_hull_explanation}A, we compute the correspondent attainable wrench-sequences for the sampled interior and edge pressures shown in Fig. \ref{fig:wrench_hull_explanation}B. These examples illustrate that the wrench-hull properties utilized in our simplified problem formulation hold across a wide variety of shapes. Note that both the direction and magnitude of the shift in each wrench-space along $i$ corresponds to sign and severity of the change in curvature along the arm centerline shape.}
    \label{fig:wrench_hull_examples}
    \vspace{-1.25em}
\end{figure*}

It is difficult to directly determine if $\mathbf{w}^*$ is in $\mathbf{a}[\mathcal{P}]$, as $\mathbf{a}[\mathcal{P}]$ is high dimensional and we lack insights into its structure. However, we have found two properties that are shared by all elements in $\mathbf{a}[\mathcal{P}]$ and are illustrated in Fig. \ref{fig:wrench_hull_examples} across 150 sample points. Checking if $\mathbf{w}^*$ satisfies these two properties serves as a close-to-sufficient test for whether it lies in $\mathbf{a}[\mathcal{P}]$.

\subsubsection{Absolute Attainability} Our first condition is to check that each absolute requirement wrench $\mathbf{w}_i^*$ lies within each $\mathbf{a}_i[\mathcal{P}]$, which we call the "$i$-th attainable wrench space". We claim the following propositions for computing each $\mathbf{a}_i[\mathcal{P}]$:


\begin{itemize}
    \item{Each $\mathbf{a}_i[\mathcal{P}]$ is convex.}
    \item{The boundary of each $\mathbf{a}_i[\mathcal{P}]$ is a subset of the image of the pressure space boundary $\partial \mathcal{P}$, i.e., $\partial \mathbf{a}_i[\mathcal{P}] \subset \mathbf{a}_i[\partial \mathcal{P}]$ }
\end{itemize}

These propositions are consequences of our actuator models being continuous and monotonic, which preserve topological invariants like boundaries and connectedness \cite{munkres_topology_2000} (see SI for proof). We can therefore approximate whether $\mathbf{w}^*_i$ is in $\mathbf{a}_i[\mathcal{P}]$ by sampling a set of points $\mathbf{E}$ from the edges of $\partial \mathcal{P}$ and checking if $\mathbf{w}^*_i$ is enclosed by the convex hull $\textrm{Conv}(\mathbf{a}_i[{\mathbf{E}}])$. We call this $\mathcal{H}_i$, the \textit{\textbf{$\mathit{\mathbf{i}}$-th absolute wrench hull}}, which also visualizes the possible reactions across $\mathcal{P}$. Across all $i$, we will refer to the collection $\mathcal{H} = (\mathcal{H}_1, \dots, \mathcal{H}_N)$ as the \textit{\textbf{absolute wrench hull}}, and if each $\mathbf{w}_i^*$ is enclosed in each $\mathcal{H}_i$ then we say the task $({}_{o}\rightgroupderiv{g}, \mathbf{q}_{\textrm{tip}}^*)$ is \textit{\textbf{absolutely attainable}} (Fig. \ref{fig:wrench_hull_explanation}D). If any $\mathbf{w}^*_i$ are not in $\mathcal{H}_i$, the shortest distance between them can be found by solving a convex quadratic program described in our SI, and we will refer to their sum as the \textbf{\textit{absolute unattainability}}.

\subsubsection{Relative Attainability} Absolute attainability is insufficient alone to guarantee that a task is truly attainable. If a requirement wrench sequence does not change along $i$ in the same "direction" as \textit{any} attainable wrench sequences - as in Fig. \ref{fig:wrench_hull_explanation}D - then it cannot be truly attainable. We now formalize this "direction" as a second test for attainability.

Consider the difference $\Delta \mathbf{w}_i^*$ of each $\mathbf{w}_i^*$ from its starting point $\mathbf{w}_1^*$. We will refer to the sequence of such differences i.e., $\Delta \mathbf{w}^* = (\mathbf{0}, \mathbf{w}_2^* - \mathbf{w}_1^*, \ldots, \mathbf{w}_N^* - \mathbf{w}_1^*)$, as a \textit{\textbf{relative requirement wrench sequence}}. For any attainable wrench sequence $\mathbf{a}(\mathbf{p})$, a similar sequence $\Delta \mathbf{a}(\mathbf{p})$ can also be defined, and we call it the \textit{\textbf{relative attainable wrench sequence}}. The collection of such sequences over $\mathcal{P}$ is the space of relative attainable wrench sequences, and is defined as:

\begin{equation}
\begin{split}
\Delta \mathbf{a}[\mathcal{P}] = \{(\mathbf{0}, \mathbf{a}_2(\mathbf{p}) - \mathbf{a}_1(\mathbf{p}), \ldots, \\
\mathbf{a}_n(\mathbf{p}) - \mathbf{a}_1(\mathbf{p})) \mid \mathbf{p} \in \mathcal{P} \}
\end{split}
\end{equation}

For a task to be attainable, there must exist $\mathbf{p}$ so that $\mathbf{a}(\mathbf{p}) = \mathbf{w}^*$, which implies that $\Delta \mathbf{a}(\mathbf{p}) = \Delta \mathbf{w}^*$. Determining if such a $\Delta \mathbf{a}(\mathbf{p})$ exists is equivalent to checking if $\Delta \mathbf{w}^*$ lies within $\Delta \mathbf{a}[\mathcal{P}]$, which is difficult. We will again approximate the problem by individually considering if each $\Delta \mathbf{w}_i^*$ lies within its respective $\Delta \mathbf{a}_i[\mathcal{P}] = \{\mathbf{a}_i(\mathbf{p}) - \mathbf{a}_1(\mathbf{p} \mid \mathbf{p} \in \mathcal{P} \}$. As with the absolute attainability, we claim the following properties of each $\Delta \mathbf{a}_i[\mathcal{P}]$:

\begin{itemize}
    \item{Each $\Delta \mathbf{a}_i[\mathcal{P}]$ is convex}
    \item{The boundary of each $\Delta \mathbf{a}_i[\mathcal{P}]$ subsets the image of the pressure space boundary $\partial \mathcal{P}$, i.e., $\partial \Delta \mathbf{a}_i[\mathcal{P}] \subset \Delta \mathbf{a}_i[\partial \mathcal{P}]$}
\end{itemize}

An extensive search yielded no counterexamples to these two properties, and their proof will be provided in a future work (see demonstrations in Fig. \ref{fig:wrench_hull_explanation}E and \ref{fig:wrench_hull_examples}). As before, we can approximate whether $\Delta \mathbf{w}_i^*$ is in $\Delta \mathbf{a}_i[\mathcal{P}]$ by checking if $\Delta \mathbf{w}_i^*$ is enclosed by $\textrm{Conv}(\Delta \mathbf{a}_i[\mathbf{E}])$, which we name the $i$-th relative wrench hull. If true for all $i$, then the task is \textit{\textbf{relatively attainable}}. If any $\Delta \mathbf{w}_i^*$ are not in $\textrm{Conv}(\Delta \mathbf{a}_i[\mathbf{E}])$, the shortest distance between them can be found by solving a convex quadratic program, and we will refer to their sum as the \textit{\textbf{relative unattainability}}.

Together, these two properties approximate whether a proposed arm design can complete a specified task. Although neither condition is sufficient to imply true attainability, when combined they are stringent enough that any task that satisfies both is functionally attainable.

\begin{figure*}[!ht]
    \centering
    \includegraphics[width=\textwidth]{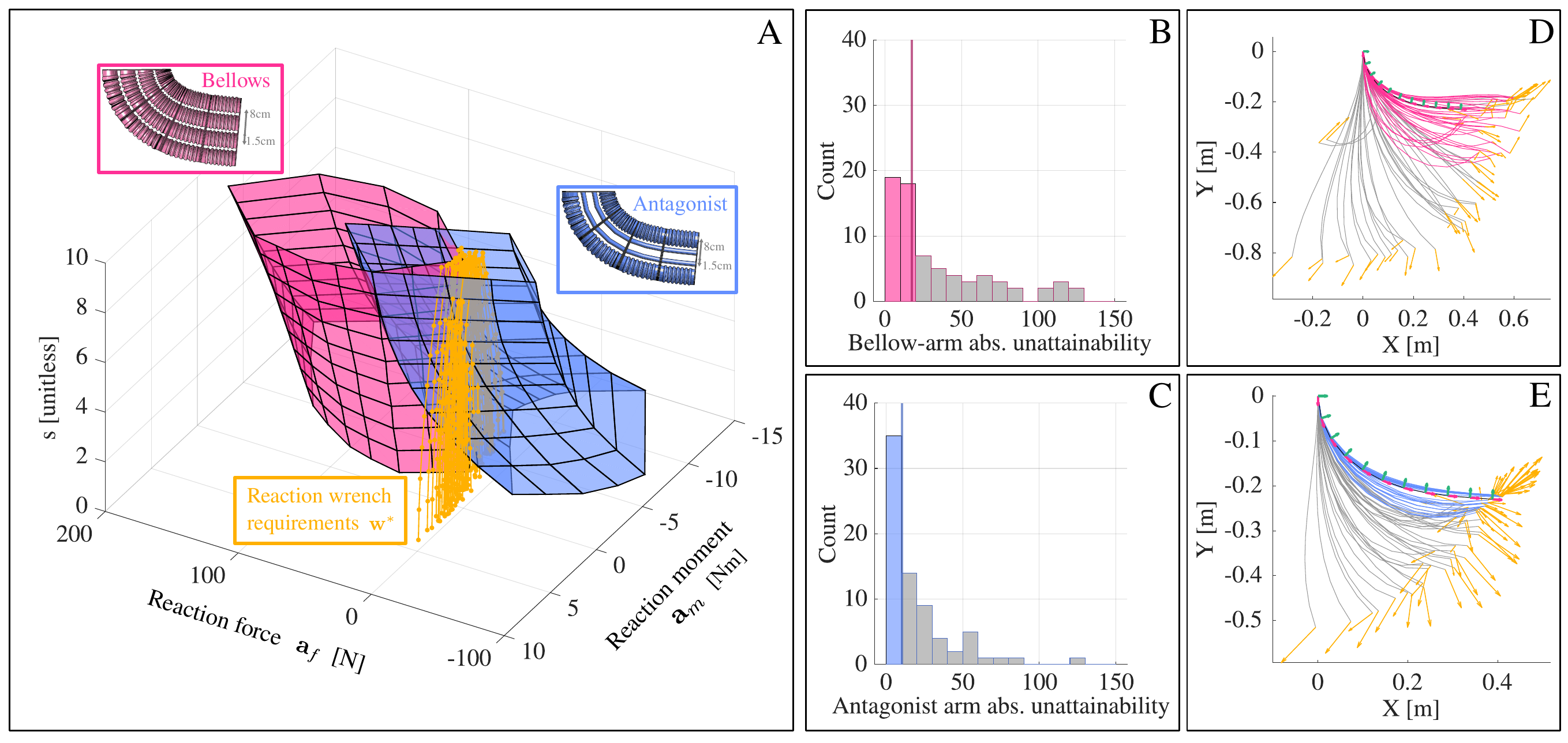}
    \caption{\textbf{Attainability methods accurately predict the ability of arm designs to accomplish a range of shared tasks}. Reaction wrench requirements are imposed by 67 sampled tip-loads applied to two arm designs held at a task shape. The attainable wrench hull of an antagonistic arm encloses significantly more wrench requirements than that of an arm with only bellows, and this is quantified in \textit{(B)} and \textit{(C)} - vertical lines are medians. Attainability thus predicts that the antagonistic arm can complete more of the tasks, and this is confirmed by exhaustively searching in \textit{(D)} and \textit{(E)} for each arm's best attempt to match the specified shape while subject to each of the sampled tip loads . Tip loads are plotted without moments, and were uniformly sampled from the range of $\begin{bmatrix} \pm 10\textrm{N} &\pm 10\textrm{N} & \pm 1\textrm{Nm}\end{bmatrix}$.}
    \label{fig:enclosure_demo}
    \vspace{-1.25em}
\end{figure*}

\section{APPLICATION TO MODEL BASED DESIGN} \label{sec:results}

We now demonstrate use of our wrench-hull method in soft robot arm design for specified tasks. Please note that \textit{blank slate} model-based design, or a rigorous, mathematical design procedure for translating a few requirements into a complete specification of materials and geometry, is a large and open challenge. Our method is focused on a substep of this process: informative evaluation of a proposed design.

In this section, we will briefly review actuator characterization. We will validate our wrench-hull method through comparison to solution-through-search. We then compare antagonistic and non-antagonistic designs, show how antagonistic arms outperform the other two, and offer a novel interpretation of the underlying mechanics. Finally, we demonstrate potential use and discuss implications of the wrench-hull analysis to design and target task shapes. 


\subsection{Experimental Characterization of Actuators}
The proposed arm designs analyzed below are actuated by either McKibben actuators, bellows actuators or both. The force function $f(\epsilon, p)$ of the McKibben actuator is taken from \cite{olson_eulerbernoulli_2020}, while the force function of the bellows actuator is measured via a similar process, but is novel to this work - see our SI for details. The bending moment function $\tau(\kappa, p)$ is adapted from findings in \cite{garbulinski_bending_2022} - we assume a linear stiffness that varies with pressure, and use the following function:

\begin{equation}
    \tau(\kappa, p) = K \frac{p}{\bar{p}} \kappa
\end{equation}

While actuator force models were extracted from available data, bending stiffness parameters were estimated due to lack of suitable test data. We selected $K = -0.285\ \mathrm{Nm^2}$ for both McKibbens and bellows, and use $\bar{p} = 50 \mathrm{kPa}$ for bellows and $100 \mathrm{kPa}$ for McKibbens. The presented method is valid for any bending stiffness model that meets the listed assumptions, and further work is needed to identify a robust, generalizable bending stiffness model for soft arms.

\begin{figure*}[!ht]
    \centering
    \includegraphics [width=\textwidth]{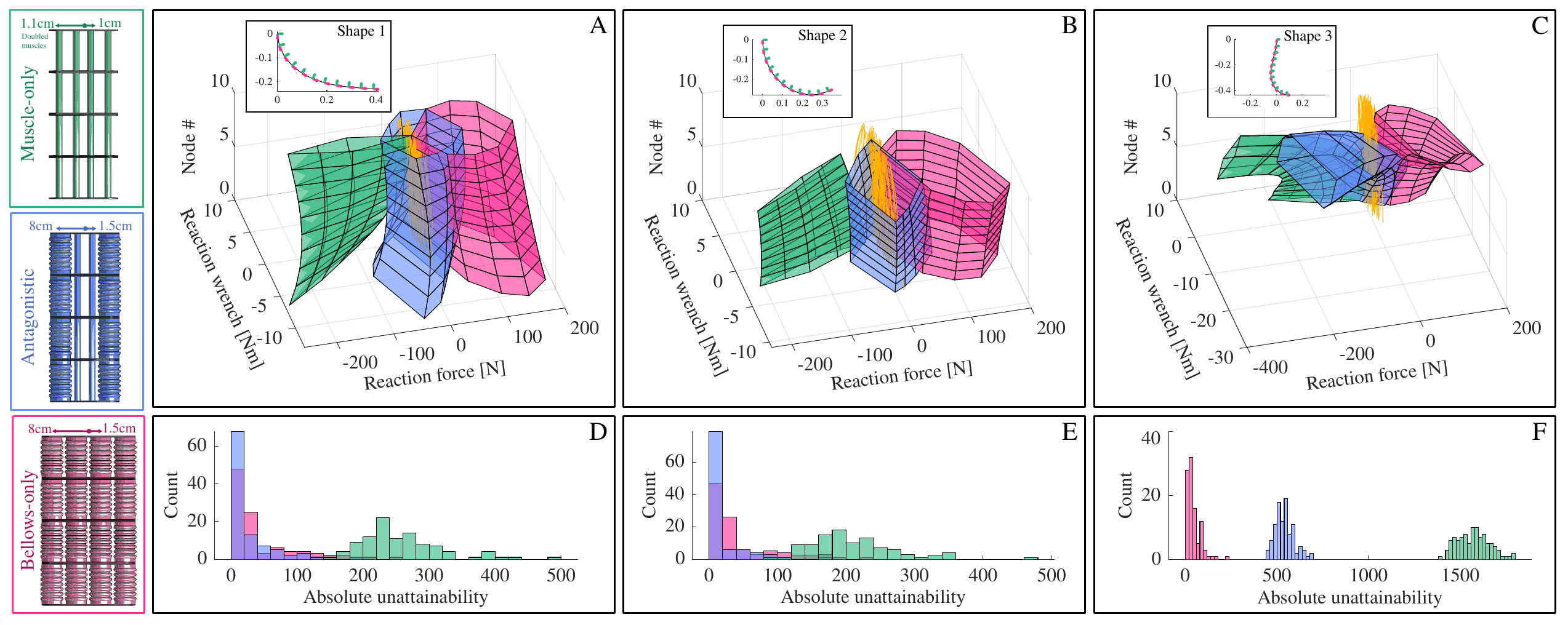}
    \caption{\textbf{Antagonistic arm designs consistently accomplish the same tasks at a wider variety of tip loads than non-antagonistic arm designs.} Across multiple task-shapes, the attainable wrench hull of the antagonistic arm design (blue) consistently encloses more internal wrench requirements imposed by different tip load wrenches than the bellows-only (red) and muscle-only (green) counterparts. In the task space, this translates to being able to attain the task-shape more for a wider range of loads, such as for shape 1 and 2. Shape 3 is more attainable for the bellows-only arm as the high curvature near the tip requires strains too large for artificial muscles to reach. }
    \label{fig:antagonism_is_better}
    \vspace{-1.25em}
\end{figure*}

\subsection{Validation of Wrench-Hull Analysis}


We consider the problem of comparing two proposed arm designs - an antagonistic arm using two McKibbens and two bellows, and a non-antagonistic arm using only bellows - over their ability to sustain a high-reaching task-shape while subject to 67 different task loads uniformly sampled from the wrench-space. The two arms' designs and dimensions are shown in Fig. \ref{fig:enclosure_demo}A, while the task-shape and force-loads are shown in Fig. \ref{fig:enclosure_demo}D and E. Using our wrench-hull methods, we are able to see that the antagonistic arm can complete a larger number of the 67 different tasks. Plotting the requirement wrench sequences induced by each of the tasks (Fig. \ref{fig:enclosure_demo}A, yellow) against the arm's absolute wrench hull $\mathcal{H}$ shows that the antagonistic arm's wrench hull encloses many more of the requirement wrench sequences than the bellows-only arm. Computing the absolute unattainability of the 67 tasks for each arm confirms that the median absolute unattainability of the tasks is much lower for the antagonistic arm. 

To confirm that tasks assigned a low absolute unattainability for a given arm are indeed more feasible for that arm, we used the search-based method described in section \ref{sec:naive_method} to find each arm's best attempt to match the task-shape while subject to each of the 67 task-loads. The resulting equilibrium arm shapes are shown in Fig. \ref{fig:enclosure_demo}D and E, with each shape's correspondent task-loads shown in yellow. As predicted by the absolute unattainability metric, the antagonistic arm's equilibrium shapes have less variance than the bellows-only arm thanks to its ability to increase stiffness to withstand larger loads. Furthermore, highlighting each arm's equilibrium shapes for task-loads that have below-median unattainability shows that, without exception, the tasks with lower unattainability are closer to being accomplished.

This test demonstrates that our wrench-hull methods accurately compares the ability of proposed arm designs to complete specified tasks. Furthermore, computing the attainability metrics took only 108 seconds, while solving for the equilibrium shapes using the search-based method took 8766 seconds, which represents a 81x speed increase.

\subsection{Evaluation of Antagonistic Actuation}

Antagonistic arms have been demonstrated to complete tasks that are difficult for McKibben- or bellows-only arms, but direct comparison of the task-competence of these arm designs has been difficult, especially over multiple tasks. In Fig. \ref{fig:antagonism_is_better}, we compare the wrench-hulls and absolute unattainability of an antagonistic, bellows-only, and McKibben-only arm over three task-shapes, each subject to 100 different task-loads. The task-shapes are each defined as a function of the node index ${}_{o}\rightgroupderiv{g}^*(i)$ and are shown in the figure.

The antagonistic arm has the lowest absolute unattainability for the first two task-shapes, followed by the bellows-only arm, and finally the McKibben-only arm. Comparing the absolute wrench-hulls of each arm suggests why the antagonistic arm is more capable of the two: because the antagonistic arm combines extending and contracting actuators, its absolute wrench-hull is closer to the origin, and therefore better at enclosing requirement wrench sequences, which are centered around the origin. The bellows-only arm may have a larger wrench-hull, but because it only has extending actuators, its wrench-hull only extends to the positive reaction force half-plane. Likewise the McKibben-only arm, whose actuators can only contract, has a wrench-hull entirely with negative reaction-force. Passive actuator reaction forces do provide countering force, as noted in \cite{me_mocap}, but are strain-dependent and small in magnitude, while antagonistic countering forces are controllable via pressure and can be large in magnitude.

On the third shape, the bellows-only arm has the lowest absolute unattainability, with the antagonistic arm a distant second. This is because the large curvature near the tip of the arm is difficult to attain with an antagonistic arm, which has higher stiffness. The bellows-only arm has lower stiffness (see Fig. \ref{fig:enclosure_demo}D) and thus can better match the shape. 

\begin{figure*}[!ht]
    \centering
    \includegraphics[width=\textwidth]{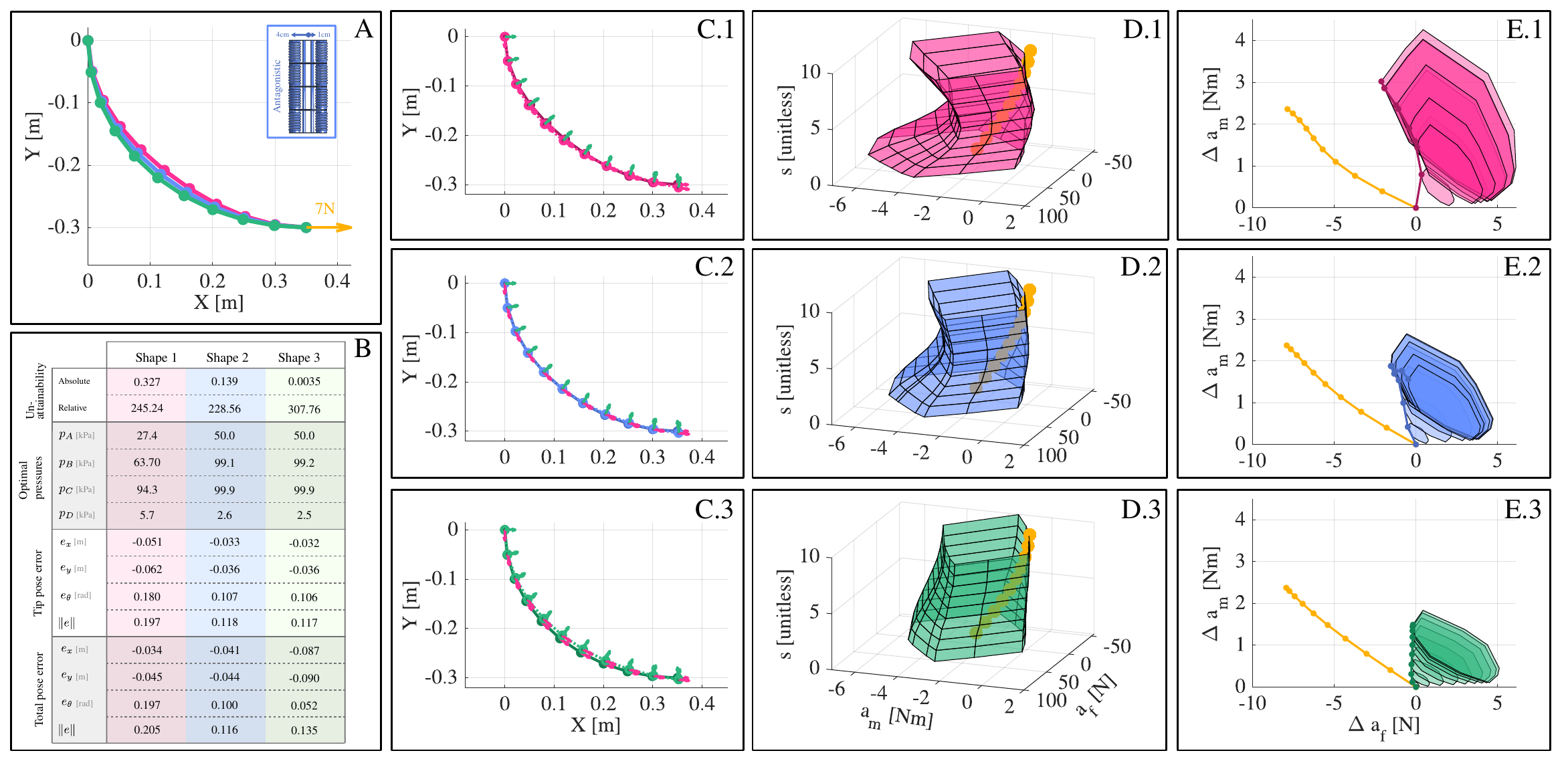}
    \caption{\textbf{Attainability methods enable direct comparison of task-shape feasibilities without simulation.} A: Three different target base-curve shapes for reaching the same tip-pose requirement, with a tip-load of 8N in positive X. C1-3: For each task-shape (solid, dark), the closest possible back-bone shape in loaded static equilibrium. equilibrium. D1-3: The absolute attainable wrench hull for each arm. E1-3: the relative attainable wrench hull for each arm, with required relative wrench superimposed in yellow.}
    \label{fig:shape_planning}
    \vspace{-1.25em}
\end{figure*}

\subsection{Evaluation of Task Shapes}

Although we have only considered design problems with requirements on the shape of the entire arm, real tasks often only place requirements on the arm's tip's pose. There are an infinite number of curves that satisfy the geometry of such tasks, but we currently lack methods to identify the smaller set of shapes that an arm can feasibly attain. This challenge deserves comprehensive consideration in future work - here we demonstrate how our wrench-hull methods to compare different arm shapes for reaching the same tip pose under the same load (Figure \ref{fig:shape_planning}). Three different task-shapes for reaching the same tip-pose are shown in Fig. \ref{fig:shape_planning} - see our SI for details. We consider a tip-load of $\begin{bmatrix} 7 & 0 & 0\end{bmatrix}^\textrm{T}$, equivalent to opening a heavy drawer.

For each task-shape, we compare the attainability metrics predicted from our wrench-hull methods to the closest arm shapes found via the search-based method. Inspecting Fig. \ref{fig:shape_planning}C suggests that the arm can best match shape 2, and this is confirmed by the total pose error metrics in Fig. \ref{fig:shape_planning}. The attainability metrics tell a similar, but more complicated story. Shape 3 actually has lower absolute unattainability than shape 2, but its relative unattainability is higher, suggesting that while shape 3's requirement wrenchs are individually more attainable, the requirement sequence as a whole is more difficult to match everywhere along the arm. 

Our results suggest fundamental limitations to shape matching caused by \textit{shared pressures}. Arms considered here are effectively single segment, but similar problems may be expected in arms with only few segments. Future work may wish to evaluate limitations to shape accuracy versus tip accuracy as a function of the number of arm segments. 


\section{CONCLUSION} \label{sec:conclusions}
In this work, we motivated and defined the problem of determining if a soft robot arm can accomplish desired tasks. We introduced a novel method to determine a task's attainability, and when a task is not attainable, we proposed new metrics to quantify the unattainability. We demonstrated our method's speed and interpretability across common arm design problems, and yielded concrete design insights by applying the method to antagonistic soft robot arms.

Although our work accelerates a key bottleneck that prevents researchers from designing soft robot arms to complete currently challenging tasks, future work is needed to develop requirement-driven-design methodologies that capitalize on the speed of our method. We could also explore selecting target shapes for efficient movement, or formalize the effects of increased pressure segmentation. Finally, while the authors acknowledge that extending this method into dynamic systems is necessary future work, there are also still many interesting results to explore in analyzing static requirements.

\color{black}



\bibliographystyle{ieeetr}
\bibliography{zotero,GinaRef_14Feb2019}
\end{document}